%% file: samplepaper.tex
\documentclass[runningheads]{llncs}
\usepackage[T1]{fontenc}
\usepackage{graphicx}
\usepackage{booktabs}
\usepackage{tikz}
\usepackage{hyperref}

\begin{document}
\title{Assessing Reliability of Symbol Detection in Concept Bottleneck Models}
%
%
\author{Javier Fumanal-Idocin\inst{1}\orcidID{0000-0002-0644-1355} \and
Javier Andreu-Perez\inst{2,3}\orcidID{0000-0002-7421-4808}}
\authorrunning{J. Fumanal-Idocin et al.}
%
\institute{University of Essex, Wivenhoe Park, Essex, United Kingdom
\email{\{j.fumanal-idocin,j.andreu-perez\}@essex.ac.uk}}
\maketitle              
\begin{abstract}
Concept Bottleneck Models (CBMs) are a relevant tool for explainable Artificial Intelligence because they make their predictions through human-interpretable symbols. However, high task accuracy does not guarantee that these symbols are detected faithfully: jointly trained CBMs may encode task-specific shortcuts in the bottleneck, making their explanations unreliable. In this paper, we study concept-detection reliability by swapping independently trained concept detectors and classification heads that share the same symbolic vocabulary. We use the resulting performance degradation, concept-level metrics, and symbol-wise uncertainty estimates to identify concepts that are especially prone to spurious firing. Finally, we propose a reliability-aware training strategy in which a shared concept detector is optimized with multiple classification heads and penalized for relying on globally or instance-wise unreliable symbols. On CUB-200-2011 with full concept supervision, detectors and heads are almost freely interchangeable (swap drop below one accuracy point, relative retention above $99\%$, and no concept detected below chance), whereas on a controlled synthetic task we show that, as the concept-supervision weight is reduced, models keep near-perfect task accuracy while swapped accuracy and agreement with the ground-truth concepts collapse to chance. Our reliability-aware training substantially mitigates this leakage, roughly doubling swap accuracy in the leaky regime.

\end{abstract}
\section{Introduction}
Explainable Models have become a reliable tool for AI deployment in real life scenarios, like financial decisions, medical settings and safety protocols. One of the most popular branches of explainable Artificial Intelligence (XAI) are neuro-symbolic methods, in which the explainability of symbolic AI is mixed with the performance of neural models \cite{garcez2023neurosymbolic}. Concept Bottleneck Models (CBMs) are popular neuro-symbolic methods which consist of two steps: first, a set of pre-established human-understandable symbols is extracted from the data; then, a light interpretable classification head is used to obtain interpretable classifications out of the symbols \cite{koh2020concept}.

There have been different approaches beyond the original CBM paper to extend their capabilities. Concept Embedding Models extend CBMs by using concept vectors instead of binary concept activations \cite{zarlenga2022concept}. Other authors have focused on the classification head: Causal Concept Graph Models extend the original linear model with a graph structure that can explicitly assess causal interactions in the reasoning process \cite{dominici2025causal}, Deep Concept Reasoners use differentiable rule-based reasoning over concept truth degrees with a fixed logic structure \cite{barbiero2023interpretable} and rule-based learning over the extracted symbols have also been proposed \cite{fumanal2023artxai}. Post-hoc Concept Bottleneck Models use CLIP or other embedding methods to identify concepts from a latent space \cite{radford2021learning,yuksekgonul2023posthoc,oikarinen2023labelfree}. Lately, leakage-aware CBMs have focused on finding reliable concepts that do not encode unintended task information, which is especially relevant when adding unsupervised concepts into the CBM \cite{havasi2022addressing}.

The reliability of the concept detection is in fact one of the main reasons for using a CBM instead of a more performant purely neural approach; however, most CBM approaches recur to joint training of predictor and concept extractor, which is prone to leaking information in the symbols. In this work, we propose a stress test for symbol reliability based on swapping independently trained concept detectors and classification heads, together with symbol-level and instance-level diagnostics that identify when individual predictions rely on spurious detections. Then, we propose a solution to this problem by training each concept detector with different head classifiers.

\section{Related Work}

\paragraph{Concept Bottleneck Models.}
Concept Bottleneck Models were introduced as an intrinsically interpretable alternative to black-box neural classifiers, where predictions are forced to pass through a human-understandable concept layer \cite{koh2020concept}. This structure enables concept inspection and intervention, but its reliability depends on the assumption that the learned bottleneck represents the intended concepts rather than task-specific shortcuts. Later work has expanded this formulation in several directions. Concept Embedding Models represent each concept with a vector instead of a single scalar activation, improving expressivity while preserving a concept-level interface \cite{zarlenga2022concept}. Other approaches modify the reasoning layer after the bottleneck, for example by using symbolic rules \cite{barbiero2023interpretable,fumanal2023artxai} or causal concept graphs \cite{dominici2025causal}. A separate line of work reduces the need for dense concept labels by building post-hoc or label-free bottlenecks from pretrained vision-language representations such as CLIP \cite{radford2021learning,yuksekgonul2023posthoc,oikarinen2023labelfree}.

A central limitation of these methods is that high task performance does not necessarily imply faithful concept detection. In particular, jointly trained CBMs can leak information through the bottleneck, allowing concept activations to encode label-relevant signals that are not part of the intended concept semantics \cite{margeloiu2021concept,mahinpei2021promises,almudevar2025there}. This issue motivates our focus on concept detection reliability. Rather than only evaluating if the final classifier is accurate, we ask whether concept detectors remain meaningful when paired with other trained heads that expect the same symbolic vocabulary. A large degradation after swapping suggests that the detector and head have co-adapted to spurious concept encodings and information leakage rather than in reliable symbol detection.

\paragraph{Uncertainty Quantification.}
Uncertainty quantification studies when a model should be trusted and what kind of uncertainty explains its errors. A common distinction is between aleatoric uncertainty, which comes from irreducible ambiguity in the data, and epistemic uncertainty, which comes from limited knowledge of the model or training set \cite{hullermeier2021aleatoric,wimmer2023quantifying}. Neural approaches estimate these quantities using techniques such as Monte Carlo dropout \cite{gal2016dropout}, ensembles \cite{rahaman2021uncertainty,wenzel2020hyperparameter}, calibration methods \cite{guo2017calibration}, or deterministic uncertainty-aware architectures \cite{liu2020simple,mukhoti2023deep}. These methods are typically evaluated at the level of the final predictive calibration and under distribution shift detection \cite{ovadia2019can}.

Our setting is complementary to standard uncertainty analysis: even if a CBM is well calibrated at the label level, its explanations can still be unreliable if the intermediate concepts are spuriously detected. We therefore treat concept misalignment as a form of symbolic uncertainty. The relevant question is not only if the model is uncertain about $y$, but also if the firing strength of a concept $c_k$ is stable under changes to the classifier head or if the firing strength itself is reliable or an artefact of the training model.

\section{Measuring Misalignment in CBMs}

\subsection{Spurious detections in a CBM}

The most straightforward approach for CBMs training follows a two-stage pipeline. Given a dataset $\mathcal{D}=\{(x_i,\mathbf{c}_i,y_i)\}_{i=1}^N$, where $x_i$ is an input, $\mathbf{c}_i \in \{0,1\}^K$ is the vector of annotated concepts and $y_i$ is the target label, a concept detector $g_\theta: \mathcal{X}\rightarrow\mathcal{C}$ is first trained to recover the concept vector from the input:

\begin{equation} \label{eq:concept_loss}
    \theta^* = \arg\min_\theta \sum_{i=1}^{N}\mathcal{L}_{c}\left(g_\theta(x_i),\mathbf{c}_i\right).
\end{equation}

Then, an interpretable prediction head $f_\phi: \mathcal{C}\rightarrow\mathcal{Y}$ is trained to map concepts into labels, either using the ground-truth concepts or the predictions produced by the frozen concept detector:

\begin{equation}
    \phi^* = \arg\min_\phi \sum_{i=1}^{N}\mathcal{L}_{y}\left(f_\phi(\mathbf{c}_i),y_i\right).
\end{equation}

At inference time, the complete model composes both modules as $\hat{y}=f_{\phi^*}(g_{\theta^*}(x))$. This decomposition is what makes the model interpretable: decisions are mediated by the intermediate symbolic representation $\hat{\mathbf{c}}=g_{\theta^*}(x)$, which can be inspected, validated and intervened on. Both symbol detection and classification head can be jointly trained together, which is usually done to boost performance of the system overall.

However, this pipeline tends to leak information in the concepts. This is because in the sequential setting, the detector is optimized only to match the annotated concepts, following Eq. (\ref{eq:concept_loss}). However, joint training optimizes the detector with respect to both concept recovery and downstream classification:
\begin{equation} \label{eq:joint}
    \mathcal{J}_{\mathrm{joint}}(\theta,\phi)
    = \frac{1}{N}\sum_{i=1}^{N}\left[
    \mathcal{L}_{y}\left(f_\phi(g_\theta(x_i)),y_i\right)
    + \lambda\mathcal{L}_{c}\left(g_\theta(x_i),\mathbf{c}_i\right)
    \right],
\end{equation}
where $\lambda\geq 0$ controls the strength of the concept-supervision regularizer.
When $\lambda$ is small, the detector is mostly shaped by the downstream classifier, which can incentivize concept activations to include spurious information that is useful for classification but not part of the intended concept. This results in unreliable firing strengths that may be artificially high because of other predictive information that triggers the concept.

This problem raises a natural question: why not always train only against the labelled concepts to prevent concept leakage? There are several reasons beyond pure performance gains. First, concept confounding can still occur when annotated concepts are strongly correlated in the dataset. For example, if ``wings'' always appear against a blue background, a detector may use the background as evidence for the concept even when trained with concept labels. Second, concept labels can be noisy, especially when concepts are obtained from weak supervision or LLM annotation, and enforcing them too strictly may propagate annotation errors into the bottleneck. Besides, some concept definitions can also be task-dependent. A concept such as ``large'', ``expensive'', or ``damaged'' may have a different meaning depending on the final task, so a fully task-agnostic detector may learn a semantically valid concept that is not the most useful one for the classifier.

\subsection{Measuring Concept Misalignment} \label{sec:misalignment}

We measure the misalignment between the intended concepts $\mathbf{c}_i$ and
the predicted concept activations $\hat{\mathbf{c}}_i=g_\theta(x_i)$ at three
complementary levels. First, at the \emph{population} level, we quantify
global symbol reliability by swapping concept detectors and classification
heads between independently trained CBMs. Second, at the \emph{symbol} level,
we ask whether individual concepts $c_k$ are statistically distinguishable,
using permutation tests on both the detector and the frozen head. Third, at
the \emph{instance} level, we estimate how reliable an individual symbol
prediction $\hat{c}_{ik}$ is by decomposing the uncertainty of an ensemble of
concept detectors into aleatoric and epistemic components.

\paragraph{Global symbol reliability.}
We train $M$ CBMs independently. The $m$-th model is written as the pair $(g_{\theta_m}^{(m)},f_{\phi_m}^{(m)})$, where $g_{\theta_m}^{(m)}:\mathcal{X}\rightarrow\mathcal{C}$ is the concept detector and $f_{\phi_m}^{(m)}:\mathcal{C}\rightarrow\mathcal{Y}$ is the classification head. For a detector $b$ and a head $a$, we define the swapped model
\begin{equation}
    h_{b\rightarrow a}(x)
    = f_{\phi_a}^{(a)}\left(g_{\theta_b}^{(b)}(x)\right).
\end{equation}
The diagonal model $h_{a\rightarrow a}$ is the original CBM, while each
off-diagonal model $h_{b\rightarrow a}$ with $b\neq a$ replaces the detector
with one learned independently. For every pair $(b,a)$, we report the original
accuracy $\mathrm{Acc}_{a\rightarrow a}$, the swapped accuracy
$\mathrm{Acc}_{b\rightarrow a}$, the swap drop
\begin{equation}
    \Delta_{b\rightarrow a}
    = \mathrm{Acc}_{a\rightarrow a}-\mathrm{Acc}_{b\rightarrow a},
\end{equation}
and the relative retention
\begin{equation}
    \rho_{b\rightarrow a}
    = \frac{\mathrm{Acc}_{b\rightarrow a}}{\mathrm{Acc}_{a\rightarrow a}}.
\end{equation}
A high swapped accuracy is evidence of a shared symbolic representation if $\Delta_{b\rightarrow a}$ is small and $\rho_{b\rightarrow a}$ is close to one. We also show each detector's accuracy according to the ground truth labelled concept vectors $\mathbf{c}_i$, so that swap degradation can also be separated into poor concept detection and detector--head incompatibility. A symbol is globally reliable only if detectors that claim to predict the same concept remain useful across heads other than their original one.

\paragraph{Symbol interchangeability.} It can  happen that the number of concepts fired is also a signal that the model is using. This allows reasoning like: ``if many concepts are detected, then it must be this class because this is usually the class where most concepts fire''. This kind of reasoning is also unreliable and makes the weights of the concepts deceiving. To test this error, we proposed different tests. At the head level, we test whether the frozen head genuinely relies on \emph{which} symbol fires. For each head $f_{\phi_a}^{(a)}$, we compare its observed accuracy on the oracle concept vectors $\mathbf{c}_i$ against two null distributions. The first is a concept-dimension permutation null, which randomly permutes the assignment of concept dimensions before feeding them to the head. This destroys the symbol-to-head correspondence while preserving each concept's marginal statistics. The second is a within-concept sample-shuffle null, which permutes the values of each concept across samples independently. This preserves concept prevalence but breaks the per-example coupling between symbols. We report right-tailed empirical $p$-values for both tests. If permuting symbol identities does not degrade head accuracy, the symbols are effectively interchangeable for that head and are therefore not distinguishable as distinct concepts.

\subsection{Individual prediction reliability}
To characterise how trustworthy a \emph{single} prediction is, we use an ensemble of $M$ independently trained detectors $\{g_m\}$ that operate over the fixed concept interface and apply the standard decomposition of total predictive uncertainty into aleatoric and epistemic components, at two levels linked by propagation through the frozen
head.

\paragraph{Symbol-level uncertainty.}
For input $x$ and concept $k$, each detector emits a Bernoulli probability $p_{m,k}=\sigma(g_m(x))_k$. With $\bar{p}_k=\frac{1}{M}\sum_m p_{m,k}$ and binary entropy $\mathcal{H}$, the total uncertainty $\mathcal{H}(\bar{p}_k)$ splits into an aleatoric part $\frac{1}{M}\sum_m \mathcal{H}(p_{m,k})$ and an epistemic part $\mathcal{H}(\bar{p}_k)-\frac{1}{M}\sum_m\mathcal{H}(p_{m,k})$, the mutual information between the symbol and the choice of detector. The epistemic term is the disagreement among interchangeable detectors about whether symbol $k$ fired on this particular input.

\paragraph{Attributing decisions to symbols.}
To identify \emph{which} symbol drives an uncertain decision, we freeze concept $k$ to its ensemble consensus $\bar{p}_k$ across all detectors and recompute the label epistemic uncertainty. The reduction $\Delta_k = \mathrm{epistemic} - \mathrm{epistemic}^{(\mathrm{freeze}\ k)}$ attributes the decision's epistemic uncertainty to symbol $k$, answering directly whether being unsure about symbol $k$ propagates to the final prediction. We report $\Delta_k$ per concept and the correlation between per-example symbol and label epistemic uncertainty, which quantifies how faithfully symbol disagreement is transmitted to the decision. The label-level uncertainty decomposition is then the standard form \cite{hullermeier2021aleatoric}.

\section{Improving CBM Symbol Reliability}

We propose to improve CBM reliability by reducing the co-adaptation between a single concept detector and a single classification head. Instead of training one detector with one head, we train a shared detector $g_\theta$ against an ensemble of $R$ independently initialized classification heads $\{f_{\phi_r}^{(r)}\}_{r=1}^{R}$. The resulting multi-head objective is
\begin{equation} \label{eq:multi}
    \mathcal{J}_{\mathrm{multi}}(\theta,\Phi)
    = \frac{1}{N}\sum_{i=1}^{N}\left[
    \frac{1}{R}\sum_{r=1}^{R}
    \mathcal{L}_{y}\left(f_{\phi_r}^{(r)}(g_\theta(x_i)),y_i\right)
    + \lambda\mathcal{L}_{c}\left(g_\theta(x_i),\mathbf{c}_i\right)
    \right],
\end{equation}
where $\Phi=\{\phi_r\}_{r=1}^{R}$ denotes the parameters of all classification heads and $\lambda$ controls the strength of the concept-supervision term. In this formulation, the detector receives the average classification gradient from several heads rather than the gradient preferred by one particular head. A spurious concept is therefore reinforced only if it is consistently useful across multiple heads, which makes it harder for the detector to encode head-specific shortcuts as concept activations.

However, different initializations alone are not enough to keep the heads diverse: since each head is a linear classifier and the cross-entropy loss is convex in its parameters given the concept activations, heads trained on identical inputs converge towards the same solution, which would make the ensemble equivalent to a single head at convergence. To prevent this collapse, each head receives the concept activations under an independent random dropout mask: for every head and batch, a fraction $p$ of the concept activations is zeroed out (with the remaining activations rescaled accordingly). Concept dropout keeps the heads diverse throughout training and has a second beneficial effect: no head can rely on a single shortcut concept or on the total number of fired concepts, because it must remain accurate under random deletion of part of its symbolic input. If we can afford to use different concept extractors as well, we can also include estimations of epistemic uncertainty per sample symbol-wise, which also allow us to detect which concepts were more unreliable.

After this first phase, we compute a global reliability score for each symbol as its balanced detection accuracy on a held-out validation split. To avoid using unreliable concepts, we add a regularization term that discourages the classifier from relying on concepts marked as unreliable. Let $r_k\in[0,1]$ denote the global reliability score of concept $k$, and let $\mathcal{U}_{\mathrm{epi}}(x_i,k)$ denote the instance-level epistemic uncertainty of concept $k$ for sample $x_i$. We combine both sources into an unreliability weight
\begin{equation} \label{eq:omega}
    \omega_{ik} = (1-r_k) + \beta \mathcal{U}_{\mathrm{epi}}(x_i,k),
\end{equation}
where $\beta\geq 0$ controls the contribution of instance-level uncertainty. We also penalize the sensitivity of the prediction loss to unreliable concept activations. Writing $\hat{\mathbf{c}}_i=g_\theta(x_i)$ and $\hat{c}_{ik}$ for its $k$-th component, define
\begin{equation}
    S_{irk} = \left|\frac{\partial \mathcal{L}_{y}\left(f_{\phi_r}^{(r)}(\hat{\mathbf{c}}_i),y_i\right)}{\partial \hat{c}_{ik}}\right|.
\end{equation}
The final reliability-aware objective is
\begin{equation} \label{eq:rel}
    \mathcal{J}_{\mathrm{rel}}(\theta,\Phi)
    = \mathcal{J}_{\mathrm{multi}}(\theta,\Phi)
    + \eta \frac{1}{N}\sum_{i=1}^{N}\frac{1}{R}\sum_{r=1}^{R}\sum_{k=1}^{K}
    \omega_{ik} S_{irk},
\end{equation}
where $\eta\geq 0$ controls how strongly unreliable symbols are discouraged. This penalty does not remove unreliable concepts from the bottleneck; instead, it discourages the heads from basing their decisions on concepts whose global reliability is low or whose instance-level epistemic uncertainty is high.

\section{Experiments}
\subsection{Experimental Setup}

\begin{figure}[t]
  \centering
  \includegraphics[width=\textwidth]{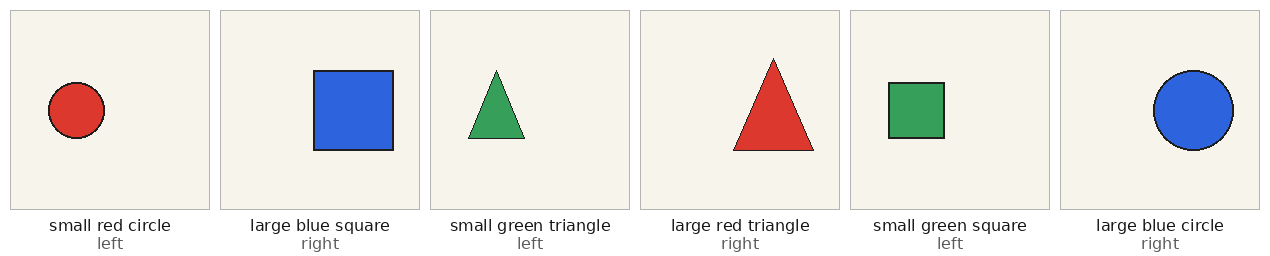}
  \caption{Example images from our synthetic coloured-shape dataset. Each scene contains one shape (circle, square, or triangle) in one of three colours (red, blue, green), drawn at one of two sizes (small, large) on the left or right of the canvas. These four attribute groups define the $10$ binary concepts of the shared vocabulary; the four five-class tasks (\texttt{shape\_color}, \texttt{shape\_position}, \texttt{color\_size}, \texttt{mixed}) assign labels through different combinations of these factors.}
  \label{fig:synthetic}
\end{figure}

  We evaluate concept-detector interchangeability on a purpose-built synthetic coloured-shape dataset (Fig.~\ref{fig:synthetic}) and CUB-200-2011~\cite{wah2011caltech}. The synthetic dataset represents shape, colour, size,
  and position using 10 binary concepts and defines four five-class tasks over this shared vocabulary. For CUB, we use all 200 bird species and the 112 binary attributes selected by the
  original CBM preprocessing, producing 4,794 training, 1,200 validation, and 5,794 test images.

  Each CBM comprises an Inception-v3 concept detector $g$ and a linear concept-to-label head $f$. The detector is initialized with ImageNet-pretrained weights and receives $299\times299$ images; its concept logits are transformed with a sigmoid before entering the head. Since our goal is to study the leakage induced by joint training, all models are trained jointly: the detector simultaneously receives the concept-supervision gradient (binary cross-entropy, weight $\lambda=1$) and the classification gradient from its head(s) (cross-entropy), following Eq.~(\ref{eq:joint}). We compare three training regimes that share this objective: (i) a \emph{joint} CBM with a single classification head, the standard formulation; (ii) the \emph{multi-head} objective of Eq.~(\ref{eq:multi}) with $R=5$ independently initialized heads, each receiving the concept activations under an independent dropout mask with $p=0.2$; and (iii) the \emph{reliability-aware} objective of Eq.~(\ref{eq:rel}), which resumes training from the multi-head checkpoint for additional epochs with the sensitivity penalty active ($\eta=0.1$, $\beta=1$). Global reliability scores $r_k$ are computed on the validation split, and instance-level epistemic uncertainty is estimated from the multi-head detectors of all seeds. For the full CUB experiments, we train five seeds per regime for 100 epochs (plus 25 reliability-aware epochs) using Adam, learning rates of $10^{-4}$ for the detector and $10^{-3}$ for the heads, and batch size 32. Swap evaluations within a regime pair the detector of one seed with the frozen head of another, so all swapped components are trained independently but under the same objective. On the synthetic dataset, we additionally sweep $\lambda\in\{0.01, 0.1, 1\}$ to quantify how concept leakage grows as the concept-supervision signal weakens relative to the classification gradient.

\subsection{Measuring Performance Degradation while Swapping Symbol Extractors and Heads}

  Table~\ref{tab:cub-swap} reports the detector--head swap on CUB-200-2011 for the three training strategies and Figure~\ref{fig:swap-heatmap} shows the full swap matrices. \emph{Original accuracy} is the species-classification accuracy of a detector paired with its own head; \emph{swapped accuracy} averages over all swapped detector--head pairings. We can see there that across all regimes the swap costs less than one accuracy point: the mean swap drop is $0.49$ points for the joint CBM, $0.84$ for the multi-head objective, and $0.64$ for the reliability-aware objective. We can also see that each head have near-perfect  performance on oracle concepts (oracle-head accuracy $\geq 98.2\%$ in every classifier), and the detectors of different seeds agree on average $95.7\%$ in their thresholded concept activations. This means that a head trained for one seed transfers almost losslessly to the detector of another. The multi-head and reliability-aware objectives do not degrade swap robustness relative to the standard joint CBM and, in fact, the multi-head proposal reaches the highest accuracy overall ($71.4\%$). 
\begin{table}[t]
  \centering
  \caption{Detector--head swap on CUB-200-2011 (200 species, 112 concepts, five seeds per regime). Entries are mean $\pm$ standard deviation over seeds; accuracies are percentages and swap drop is in accuracy points. \emph{Original} and \emph{Concept agr.} are computed over the five matched-seed pairs, \emph{Swapped} and \emph{Swap drop} over the 20 mismatched detector--head pairs. Concept agreement is the fraction of binary concept activations on which a detector agrees with the oracle.}
  \label{tab:cub-swap}
  \setlength{\tabcolsep}{4pt}
  \small
  \begin{tabular}{lccccc}
    \toprule
    Regime & Original & Swapped & Swap drop & Oracle-head & Concept agr. \\
    \midrule
    Joint              & $70.7 \pm 1.6$ & $70.2 \pm 1.8$ & $0.49 \pm 2.65$ & $99.5 \pm 0.6$ & $95.7 \pm 0.5$ \\
    Multi-head         & $71.4 \pm 0.9$ & $70.5 \pm 1.0$ & $0.84 \pm 1.53$ & $98.3 \pm 0.5$ & $95.7 \pm 0.3$ \\
    Reliability-aware  & $69.2 \pm 1.1$ & $68.6 \pm 1.3$ & $0.64 \pm 1.85$ & $98.2 \pm 0.6$ & $95.2 \pm 0.3$ \\
    \bottomrule
  \end{tabular}
\end{table}

\begin{figure}[t]
  \centering
  \resizebox{\linewidth}{!}{\input{swap_heatmap.tikz}}
  \caption{Detector--head swap-accuracy matrices on CUB-200-2011 for the three regimes. Each cell is the species accuracy (\%) of detector $g_i$ (row) paired with head $h_j$ (column); red boxes mark the matched-seed (original) pairings on the diagonal, and darker cells denote higher accuracy on a shared scale. Off-diagonal swaps are close to the diagonal in every regime: variation runs along rows (detector quality) rather than along the diagonal vs.\ off-diagonal, i.e.\ accuracy is governed by which detector is used, not by whether its head is the matched one. The single weak detector ($g_4$ for joint, top row for reliability-aware) lowers its whole row regardless of head, confirming that swap drops reflect detector strength rather than detector--head incompatibility.}
  \label{fig:swap-heatmap}
\end{figure}

  The CUB results above use full concept supervision ($\lambda=1$). To show that swap robustness is a direct consequence of that supervision rather than an inherent property of the architecture, we sweep $\lambda$ on the controlled synthetic \texttt{shape\_color} task. Table~\ref{tab:synth-lambda} reports the result. As $\lambda$ decreases, every proposal keeps near-perfect \emph{matched} accuracy, but swapped accuracy collapses from $100\%$ at $\lambda=1$ to roughly $20$--$30\%$ at $\lambda=0.01$ for the joint CBM, and the agreement between the detected concepts and the oracle falls from $98.5\%$ to $51.6\%$, i.e.\ essentially chance for binary concepts. This is leakage made explicit: high task accuracy coexists with symbols that no longer mean what they claim, so a head trained on one detector fails on another. The multi-head and reliability-aware objectives mitigate this at low $\lambda$---at $\lambda=0.1$ they roughly double the joint swap accuracy ($48$--$50\%$ vs.\ $26\%$) and raise oracle-head accuracy from $33.8\%$ to $48$--$56\%$---confirming on a controlled task the mechanism behind the proposed fix. The large standard deviations on swapped accuracy are themselves diagnostic: under leakage each seed learns its own permutation of the symbol space, so a few detector--head pairings align by chance while most do not.

\begin{table}[t]
  \centering
  \caption{Concept-supervision sweep on the synthetic \texttt{shape\_color} task (uses 10 concepts, results averaged over 5 trials). \emph{Matched} and \emph{Concept agr.} show the results for the five matched pairs, \emph{Swapped} over the 20 mismatched detector--head pairs, \emph{Oracle-head} over the five heads using the concept labels as input. As the concept-supervision weight $\lambda$ shrinks, matched accuracy stays near $100\%$ while swapped accuracy and concept--oracle agreement collapse, which shows direct evidence of concept leakage that the matched accuracy hides.}
  \label{tab:synth-lambda}
  \setlength{\tabcolsep}{4pt}
  \small
  \begin{tabular}{llcccc}
    \toprule
    $\lambda$ & Regime & Matched & Swapped & Concept agr. & Oracle-head \\
    \midrule
    $0.01$ & Joint             & $99.6 \pm 0.9$ & $20.5 \pm 24.5$ & $51.6 \pm 5.0$ & $23.3 \pm 6.3$ \\
    $0.01$ & Multi-head        & $100.0 \pm 0.0$ & $27.1 \pm 21.9$ & $52.9 \pm 6.7$ & $29.2 \pm 11.5$ \\
    $0.01$ & Reliability-aware & $100.0 \pm 0.0$ & $29.5 \pm 21.4$ & $52.9 \pm 6.6$ & $31.4 \pm 8.6$ \\
    \midrule
    $0.1$  & Joint             & $100.0 \pm 0.0$ & $26.1 \pm 24.2$ & $56.6 \pm 6.6$ & $33.8 \pm 11.0$ \\
    $0.1$  & Multi-head        & $100.0 \pm 0.0$ & $48.6 \pm 22.5$ & $62.3 \pm 7.1$ & $48.2 \pm 14.1$ \\
    $0.1$  & Reliability-aware & $100.0 \pm 0.0$ & $50.5 \pm 22.8$ & $62.3 \pm 6.8$ & $55.7 \pm 14.2$ \\
    \midrule
    $1.0$  & Joint             & $100.0 \pm 0.0$ & $100.0 \pm 0.0$ & $98.5 \pm 0.8$ & $100.0 \pm 0.0$ \\
    $1.0$  & Multi-head        & $100.0 \pm 0.0$ & $100.0 \pm 0.0$ & $99.8 \pm 0.2$ & $100.0 \pm 0.0$ \\
    $1.0$  & Reliability-aware & $100.0 \pm 0.0$ & $100.0 \pm 0.0$ & $99.4 \pm 0.9$ & $100.0 \pm 0.0$ \\
    \bottomrule
  \end{tabular}
\end{table}

\subsection{Measuring Reliable Symbol Firing}

  Table~\ref{tab:cub-reliability} summarises concept-level identifiability and the head-compatibility null tests on CUB. Concepts are detected well above chance in every regime: the mean per-concept balanced accuracy is $0.91$ for the joint and multi-head objectives and $0.90$ for the reliability-aware objective, and in all three regimes none of the 112 concepts is unreliable (balanced accuracy below $0.6$ or permutation $p>0.05$). The frozen head genuinely relies on the concept detection rather than on incidental correlations: under both the concept-dimension permutation null and the prevalence-preserving sample-shuffle null, the head collapses to near-chance accuracy ($\approx 0.5\%$ on 200 classes) against an observed accuracy of $98$--$100\%$, with permutation $p<0.002$ in every case ($1000$ permutations). The reliability-aware objective trades a small amount of mean balanced accuracy (from $0.910$ to $0.896$).

\begin{table}[t]
  \centering
  \caption{Reliable symbol firing on CUB-200-2011. Concept balanced accuracy is the mean $\pm$ std over the 112 concepts; ``Unreliable'' counts concepts with balanced accuracy $<0.6$ or permutation $p>0.05$. The head-compatibility columns give the observed frozen-head accuracy and the mean accuracy under the concept-permutation and sample-shuffle nulls ($1000$ permutations, $p<0.002$ throughout).}
  \label{tab:cub-reliability}
  \begin{tabular}{lccccc}
    \toprule
    Regime & Concept bal.\ acc. & Unreliable & Head acc. & Perm.\ null & Shuffle null \\
    \midrule
    Joint              & $0.910 \pm 0.029$ & 0/112 & 1.000 & 0.005 & 0.005 \\
    Multi-head         & $0.909 \pm 0.029$ & 0/112 & 0.980 & 0.005 & 0.005 \\
    Reliability-aware  & $0.896 \pm 0.032$ & 0/112 & 0.985 & 0.005 & 0.005 \\
    \bottomrule
  \end{tabular}
\end{table}

\subsection{Epistemic Uncertainty of symbol firing}

  Table~\ref{tab:cub-uncertainty} decomposes the ensemble uncertainty of the detector ensemble on CUB. Symbol-level epistemic uncertainty (ensemble disagreement on individual concept activations) is low and comparable across regimes ($0.032$--$0.037$), whereas the epistemic uncertainty propagated to the predicted label is an order of magnitude larger ($0.29$--$0.36$), reflecting that small disagreements over many concepts compound at the head. Uncertainty is diagnostic of error: in every regime label epistemic uncertainty is negatively correlated with per-example correctness (Spearman $\rho \approx -0.46$ to $-0.48$), and symbol-level and label-level epistemic uncertainty are strongly coupled ($\rho \approx 0.91$--$0.93$), so the concepts the ensemble is unsure about are the ones that drive uncertain predictions. The reliability-aware objective shows the highest label epistemic uncertainty ($0.356$) and the strongest uncertainty--error correlation, consistent with its lower matched accuracy.

\begin{table}[t]
  \centering
  \caption{Epistemic uncertainty of symbol firing on CUB-200-2011, estimated from the five-seed detector ensemble. ``Symbol epistemic'' and ``Label epistemic'' are mean per-example epistemic uncertainties at the concept and propagated-label levels; ``Ensemble acc.'' is the accuracy of the averaged ensemble. The last two columns are Spearman correlations of label epistemic uncertainty with per-example correctness and of symbol- with label-level epistemic uncertainty.}
  \label{tab:cub-uncertainty}
  \begin{tabular}{lccccc}
    \toprule
    Regime & Symbol epi. & Label epi. & Ensemble acc. & $\rho$(label,\,correct) & $\rho$(symbol,\,label) \\
    \midrule
    Joint              & 0.033 & 0.305 & 0.776 & $-0.462$ & 0.927 \\
    Multi-head         & 0.032 & 0.294 & 0.776 & $-0.463$ & 0.931 \\
    Reliability-aware  & 0.037 & 0.356 & 0.765 & $-0.478$ & 0.913 \\
    \bottomrule
  \end{tabular}
\end{table}

\begin{figure}[t]
  \centering
  \begin{minipage}[c]{0.27\textwidth}
    \centering
    \includegraphics[width=\linewidth]{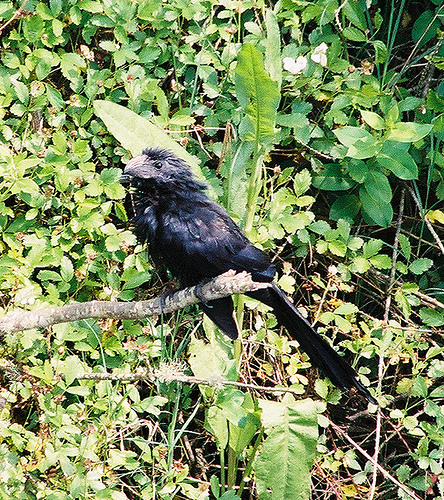}\\[2pt]
    {\footnotesize true: \emph{Groove-billed Ani}}
  \end{minipage}\hfill
  \begin{minipage}[c]{0.70\textwidth}
    \centering
    \resizebox{\linewidth}{!}{%
\begin{tikzpicture}[font=\scriptsize]
  \node at (0.460,4.380) {$g_{1}$};
  \node at (1.380,4.380) {$g_{2}$};
  \node at (2.300,4.380) {$g_{3}$};
  \node at (3.220,4.380) {$g_{4}$};
  \node at (4.140,4.380) {$g_{5}$};
  \fill[black!99] (0.000,3.600) rectangle (0.920,4.200);
  \node[text=white] at (0.460,3.900) {0.99};
  \fill[black!99] (0.920,3.600) rectangle (1.840,4.200);
  \node[text=white] at (1.380,3.900) {0.99};
  \fill[black!100] (1.840,3.600) rectangle (2.760,4.200);
  \node[text=white] at (2.300,3.900) {1.00};
  \fill[black!100] (2.760,3.600) rectangle (3.680,4.200);
  \node[text=white] at (3.220,3.900) {1.00};
  \fill[black!100] (3.680,3.600) rectangle (4.600,4.200);
  \node[text=white] at (4.140,3.900) {1.00};
  \node[anchor=east] at (-0.120,3.900) {\texttt{has\_primary\_color::black}};
  \fill[black!100] (0.000,3.000) rectangle (0.920,3.600);
  \node[text=white] at (0.460,3.300) {0.99};
  \fill[black!100] (0.920,3.000) rectangle (1.840,3.600);
  \node[text=white] at (1.380,3.300) {0.99};
  \fill[black!100] (1.840,3.000) rectangle (2.760,3.600);
  \node[text=white] at (2.300,3.300) {1.00};
  \fill[black!100] (2.760,3.000) rectangle (3.680,3.600);
  \node[text=white] at (3.220,3.300) {1.00};
  \fill[black!100] (3.680,3.000) rectangle (4.600,3.600);
  \node[text=white] at (4.140,3.300) {1.00};
  \node[anchor=east] at (-0.120,3.300) {\texttt{has\_wing\_color::black}};
  \fill[black!100] (0.000,2.400) rectangle (0.920,3.000);
  \node[text=white] at (0.460,2.700) {1.00};
  \fill[black!100] (0.920,2.400) rectangle (1.840,3.000);
  \node[text=white] at (1.380,2.700) {1.00};
  \fill[black!100] (1.840,2.400) rectangle (2.760,3.000);
  \node[text=white] at (2.300,2.700) {1.00};
  \fill[black!100] (2.760,2.400) rectangle (3.680,3.000);
  \node[text=white] at (3.220,2.700) {1.00};
  \fill[black!100] (3.680,2.400) rectangle (4.600,3.000);
  \node[text=white] at (4.140,2.700) {1.00};
  \node[anchor=east] at (-0.120,2.700) {\texttt{has\_eye\_color::black}};
  \fill[black!100] (0.000,1.800) rectangle (0.920,2.400);
  \node[text=white] at (0.460,2.100) {1.00};
  \fill[black!100] (0.920,1.800) rectangle (1.840,2.400);
  \node[text=white] at (1.380,2.100) {1.00};
  \fill[black!100] (1.840,1.800) rectangle (2.760,2.400);
  \node[text=white] at (2.300,2.100) {1.00};
  \fill[black!100] (2.760,1.800) rectangle (3.680,2.400);
  \node[text=white] at (3.220,2.100) {0.99};
  \fill[black!100] (3.680,1.800) rectangle (4.600,2.400);
  \node[text=white] at (4.140,2.100) {1.00};
  \node[anchor=east] at (-0.120,2.100) {\texttt{has\_size::small}};
  \fill[black!1] (0.000,1.200) rectangle (0.920,1.800);
  \node[text=black] at (0.460,1.500) {0.01};
  \fill[black!0] (0.920,1.200) rectangle (1.840,1.800);
  \node[text=black] at (1.380,1.500) {0.00};
  \fill[black!0] (1.840,1.200) rectangle (2.760,1.800);
  \node[text=black] at (2.300,1.500) {0.00};
  \fill[black!0] (2.760,1.200) rectangle (3.680,1.800);
  \node[text=black] at (3.220,1.500) {0.00};
  \fill[black!0] (3.680,1.200) rectangle (4.600,1.800);
  \node[text=black] at (4.140,1.500) {0.00};
  \node[anchor=east] at (-0.120,1.500) {\texttt{has\_primary\_color::yellow}};
  \fill[black!3] (0.000,0.600) rectangle (0.920,1.200);
  \node[text=black] at (0.460,0.900) {0.03};
  \fill[black!42] (0.920,0.600) rectangle (1.840,1.200);
  \node[text=black] at (1.380,0.900) {0.42};
  \fill[black!96] (1.840,0.600) rectangle (2.760,1.200);
  \node[text=white] at (2.300,0.900) {0.95};
  \fill[black!3] (2.760,0.600) rectangle (3.680,1.200);
  \node[text=black] at (3.220,0.900) {0.03};
  \fill[black!93] (3.680,0.600) rectangle (4.600,1.200);
  \node[text=white] at (4.140,0.900) {0.93};
  \node[anchor=east] at (-0.120,0.900) {\texttt{has\_bill\_shape::all-purpose}};
  \draw[red,very thick] (0,0.600) rectangle (4.600,1.200);
  \fill[black!2] (0.000,0.000) rectangle (0.920,0.600);
  \node[text=black] at (0.460,0.300) {0.02};
  \fill[black!58] (0.920,0.000) rectangle (1.840,0.600);
  \node[text=white] at (1.380,0.300) {0.58};
  \fill[black!97] (1.840,0.000) rectangle (2.760,0.600);
  \node[text=white] at (2.300,0.300) {0.97};
  \fill[black!3] (2.760,0.000) rectangle (3.680,0.600);
  \node[text=black] at (3.220,0.300) {0.03};
  \fill[black!99] (3.680,0.000) rectangle (4.600,0.600);
  \node[text=white] at (4.140,0.300) {0.99};
  \node[anchor=east] at (-0.120,0.300) {\texttt{has\_leg\_color::grey}};
  \draw[red,very thick] (0,0.000) rectangle (4.600,0.600);
  \draw[gray] (0,0) grid[xstep=0.920,ystep=0.600] (4.600,4.200);
\end{tikzpicture}}\\[3pt]
    {\footnotesize five detectors $g_1\!\dots\!g_5\;\rightarrow\;$ frozen head $f$: predicts \emph{Shiny Cowbird} ($\times$ wrong)}
  \end{minipage}
  \caption{Worked example from the joint regime. Each cell is the firing probability of a symbol under one of the five interchangeable detectors $g_1,\dots,g_5$ (darker $=$ closer to $1$). The black-plumage symbols fire near $1$ with full agreement, but the two red-boxed symbols---\texttt{has\_leg\_color::grey} and \texttt{has\_bill\_shape::all-purpose}---split the detectors. For \texttt{has\_leg\_color::grey} the firing probabilities are $0.02,0.59,0.97,0.03,0.99$ (symbol epistemic uncertainty $0.47$), and freezing it to the ensemble consensus removes $\Delta_k=0.18$ of the prediction's epistemic uncertainty. This one contested symbol flips the frozen head to a visually similar all-dark \emph{Shiny Cowbird} (cross-entropy $1.62$) instead of the true \emph{Groove-billed Ani}.}
  \label{fig:worked-example}
\end{figure}

  A concrete case illustrates how symbol disagreement propagates to a decision. On a test image of a \emph{Groove-billed Ani} (joint regime; Fig.~\ref{fig:worked-example}), the five interchangeable detectors split sharply on the symbol \texttt{has\_leg\_color::grey}, emitting firing probabilities of $0.02$, $0.59$, $0.97$, $0.03$, and $0.99$; this is almost maximal epistemic uncertainty for a single symbol ($0.47$). Freezing that symbol to the ensemble consensus removes $\Delta_k = 0.18$ of the prediction's epistemic uncertainty, which is among the largest single-symbol attributions in the run, so this one contested leg-color symbol carries a substantial share of the decision's instability. The prediction indeed fails: the model labels the bird a \emph{Shiny Cowbird}, a visually similar all-dark species, with an elevated cross-entropy of $1.62$. This example shows the how a symbol on which  detectors disagree, and which the head weights heavily, is most prone to result in erratic predictions.

\section{Discussion and Conclusions}
In this paper we have studied the reliability of concept detection in Concept Bottleneck Models (CBMs) from different perspectives. We have used global metrics to compute the reliability of concept detection, and found that indeed some concepts can be more reliably detected than others. We also used different classification heads to check how concept detection can be expected to work under classification heads other than the original one, which can also be used as a signal of how much information the concept detector is leaking. We have also performed experiments to check how common and relevant for performance epistemic uncertainty is in CBMs. Finally, we have proposed a simple fix: to train the symbol extractor with different classification heads, which mitigates the issue of information leakage.

We found in the results using out synthetic dataset that without proper supervision, information leakage is very prone to happen in symbol detection. For the CUB dataset we found that concept detection is mostly reliable, although evidence for unreliable concepts was found using epistemic uncertainty-based analysis of individual predictions. The two mitigations strategies worked to reduce the concept leakage, and the ensemble-head approach improved accuracy as well. Penalising unreliable concepts seems to mitigate concept spurious firings, but it also hurt performance in many cases. How to harmonize this penalization with high-performance aims is left as future work.

As future lines, our research should extend this analysis for other CBM architectures such as Concept Embedding Models \cite{zarlenga2022concept} and Causal Concept Bottleneck Models \cite{dominici2025causal}. We also intend to expand our analysis to Large Language Model-generated labels. 

Code is available at: \url{https://github.com/Fuminides/cbm_sanity}

\bibliographystyle{splncs04}
\bibliography{mybibliograhpy}

\end{document}

%% file: swap_heatmap.tikz
\begin{tikzpicture}[font=\scriptsize]
	\def\cell{0.62}
	
	\newcommand{\heatpanel}[3]{%
		\begin{scope}[shift={(#1,0)}]
			
			\foreach \j in {0,...,4} {
				\node at ({(\j+0.5)*\cell},{5*\cell+0.16}) {$h_\j$};
			}
			
			\node[font=\scriptsize\bfseries] at ({2.5*\cell},{5*\cell+0.52}) {#2};
			
			#3
			
			\draw[gray] (0,0) grid[xstep=\cell,ystep=\cell] ({5*\cell},{5*\cell});
			
			\foreach \i in {0,...,4} {
				\draw[red,very thick]
				({\i*\cell},{(4-\i)*\cell})
				rectangle ({(\i+1)*\cell},{(5-\i)*\cell});
			}
			
		\end{scope}
	}
	
	\newcommand{\cellentry}[5]{%
		\fill[black!#4]
		({#2*\cell},{#1*\cell})
		rectangle ({(#2+1)*\cell},{(#1+1)*\cell});
		\ifnum#5=1
		\node[text=white] at ({(#2+0.5)*\cell},{(#1+0.5)*\cell}) {#3};
		\else
		\node[text=black] at ({(#2+0.5)*\cell},{(#1+0.5)*\cell}) {#3};
		\fi
	}
	
	\foreach \i/\g in {0/g_4,1/g_3,2/g_2,3/g_1,4/g_0} {
		\node[anchor=east] at (-0.10,{(\i+0.5)*\cell}) {$\g$};
	}
	
	\heatpanel{0}{Joint}{
		\cellentry{4}{0}{71.6}{75}{1}
		\cellentry{4}{1}{71.4}{72}{1}
		\cellentry{4}{2}{71.2}{70}{1}
		\cellentry{4}{3}{71.5}{73}{1}
		\cellentry{4}{4}{71.1}{69}{1}
		
		\cellentry{3}{0}{71.3}{70}{1}
		\cellentry{3}{1}{71.6}{74}{1}
		\cellentry{3}{2}{71.2}{69}{1}
		\cellentry{3}{3}{71.0}{68}{1}
		\cellentry{3}{4}{70.8}{64}{1}
		
		\cellentry{2}{0}{71.1}{68}{1}
		\cellentry{2}{1}{71.3}{70}{1}
		\cellentry{2}{2}{71.8}{76}{1}
		\cellentry{2}{3}{71.6}{74}{1}
		\cellentry{2}{4}{71.2}{70}{1}
		
		\cellentry{1}{0}{71.1}{69}{1}
		\cellentry{1}{1}{70.6}{62}{1}
		\cellentry{1}{2}{70.9}{66}{1}
		\cellentry{1}{3}{71.2}{69}{1}
		\cellentry{1}{4}{70.8}{65}{1}
		
		\cellentry{0}{0}{66.9}{19}{0}
		\cellentry{0}{1}{66.9}{19}{0}
		\cellentry{0}{2}{66.8}{18}{0}
		\cellentry{0}{3}{66.3}{12}{0}
		\cellentry{0}{4}{67.5}{27}{0}
	}
	
	\heatpanel{3.65}{Multi-head}{
		\cellentry{4}{0}{69.8}{54}{1}
		\cellentry{4}{1}{69.3}{48}{0}
		\cellentry{4}{2}{68.6}{39}{0}
		\cellentry{4}{3}{68.7}{41}{0}
		\cellentry{4}{4}{69.2}{46}{0}
		
		\cellentry{3}{0}{70.6}{62}{1}
		\cellentry{3}{1}{71.6}{75}{1}
		\cellentry{3}{2}{70.9}{66}{1}
		\cellentry{3}{3}{71.0}{67}{1}
		\cellentry{3}{4}{70.6}{63}{1}
		
		\cellentry{2}{0}{71.8}{77}{1}
		\cellentry{2}{1}{72.3}{82}{1}
		\cellentry{2}{2}{72.4}{84}{1}
		\cellentry{2}{3}{71.4}{72}{1}
		\cellentry{2}{4}{71.8}{77}{1}
		
		\cellentry{1}{0}{71.0}{68}{1}
		\cellentry{1}{1}{70.9}{66}{1}
		\cellentry{1}{2}{70.5}{61}{1}
		\cellentry{1}{3}{71.9}{78}{1}
		\cellentry{1}{4}{70.9}{66}{1}
		
		\cellentry{0}{0}{70.8}{65}{1}
		\cellentry{0}{1}{70.1}{56}{1}
		\cellentry{0}{2}{69.9}{55}{1}
		\cellentry{0}{3}{70.4}{60}{1}
		\cellentry{0}{4}{71.1}{68}{1}
	}
	
	\heatpanel{7.30}{Reliability-aware}{
		\cellentry{4}{0}{67.5}{27}{0}
		\cellentry{4}{1}{67.2}{22}{0}
		\cellentry{4}{2}{66.6}{16}{0}
		\cellentry{4}{3}{66.3}{13}{0}
		\cellentry{4}{4}{67.2}{22}{0}
		
		\cellentry{3}{0}{68.0}{32}{0}
		\cellentry{3}{1}{69.0}{44}{0}
		\cellentry{3}{2}{68.7}{40}{0}
		\cellentry{3}{3}{68.2}{34}{0}
		\cellentry{3}{4}{68.0}{32}{0}
		
		\cellentry{2}{0}{70.1}{57}{1}
		\cellentry{2}{1}{70.5}{61}{1}
		\cellentry{2}{2}{70.4}{61}{1}
		\cellentry{2}{3}{69.9}{54}{1}
		\cellentry{2}{4}{70.1}{56}{1}
		
		\cellentry{1}{0}{68.1}{34}{0}
		\cellentry{1}{1}{68.4}{36}{0}
		\cellentry{1}{2}{67.4}{26}{0}
		\cellentry{1}{3}{68.8}{41}{0}
		\cellentry{1}{4}{67.6}{28}{0}
		
		\cellentry{0}{0}{70.0}{56}{1}
		\cellentry{0}{1}{69.6}{51}{1}
		\cellentry{0}{2}{69.7}{53}{1}
		\cellentry{0}{3}{69.7}{52}{1}
		\cellentry{0}{4}{70.2}{58}{1}
	}
	
\end{tikzpicture}